\documentclass[letterpaper,journal]{IEEEtran}
\usepackage{amsmath,amsfonts}
\usepackage{algorithm}
\usepackage{array}
\usepackage[caption=false,font=scriptsize,labelfont=sf,textfont=sf]{subfig}
\usepackage{textcomp}
\usepackage{stfloats}
\usepackage{url}
\usepackage{verbatim}
\usepackage{graphicx}

\usepackage{amsmath,amssymb}
\usepackage{algpseudocodex}
\usepackage{booktabs}
\usepackage{threeparttable}
\usepackage{enumitem}

\usepackage{siunitx}
\usepackage{cleveref}

\usepackage[backend=biber,style=ieee,natbib=true,maxcitenames=2,mincitenames=1,doi=false,isbn=false,url=false,eprint=false]{biblatex}
\begin{document}

\title{UPLME: Uncertainty-Aware Probabilistic Language Modelling for Robust Empathy Regression}
\author{Md~Rakibul~Hasan,~\IEEEmembership{Graduate~Student~Member,~IEEE,}
Md~Zakir~Hossain,~\IEEEmembership{Member,~IEEE,}
Aneesh~Krishna,
Shafin~Rahman,
and~Tom~Gedeon,~\IEEEmembership{Senior~Member,~IEEE}
\thanks{M R Hasan, M Z Hossain, A Krishna and T Gedeon are with School of Electrical Engineering, Computing and Mathematical Sciences, Curtin University, Bentley WA 6102, Australia.}
\thanks{S Rahman is with North South University, Dhaka 1229, Bangladesh.}
\thanks{M R Hasan is also with BRAC University, Dhaka 1212, Bangladesh.}
\thanks{M Z Hossain and T Gedeon are also with The Australian National University, Canberra ACT 2600, Australia.}
\thanks{T Gedeon is also with University of ÓBuda, 1034 Budapest, Hungary.}
\thanks{E-mail: \{rakibul.hasan, zakir.hossain1, a.krishna, tom.gedeon\}@curtin.edu.au, shafin.rahman@northsouth.edu}
\thanks{Corresponding author: M R Hasan}
}

\markboth{Under review}%
{Hasan \MakeLowercase{\textit{et al.}}: UPLME: Uncertainty-Aware Probabilistic Language Modelling for Robust Empathy Regression}

\IEEEpubid{This work has been submitted to the IEEE for possible publication. Copyright may be transferred without notice.}

\maketitle

\begin{abstract}
Noisy self-reported empathy scores challenge supervised learning for empathy regression. While many algorithms have been proposed for learning with noisy labels in textual \emph{classification} problems, the \emph{regression} counterpart is relatively under-explored. We propose \emph{UPLME}, an uncertainty-aware probabilistic language modelling framework to capture label noise in empathy regression tasks. One of the novelties in UPLME is a probabilistic language model that predicts both empathy scores and heteroscedastic uncertainty, and is trained using Bayesian concepts with variational model ensembling. We further introduce two novel loss components: one penalises degenerate Uncertainty Quantification (UQ), and another enforces similarity between the input pairs on which empathy is being predicted. UPLME achieves state-of-the-art performance (Pearson Correlation Coefficient: $0.558\rightarrow0.580$ and $0.629\rightarrow0.634$) in terms of the performance reported in the literature on two public benchmarks with label noise. Through synthetic label noise injection, we demonstrate that UPLME is effective in distinguishing between noisy and clean samples based on the predicted uncertainty. UPLME further outperform (Calibration error: $0.571\rightarrow0.376$) a recent variational model ensembling-based UQ method designed for regression problems. Code is publicly available at \url{https://github.com/hasan-rakibul/UPLME}.
\end{abstract}


\begin{figure}[t!]
    \centering
    \subfloat[\textbf{Top}: predicting empathy scores in written essays and an example of how label noise arises in this task: self-reported scores (an empathy score of 1.0 means the lowest empathy) did not align with the corresponding essay. \textbf{Bottom}: predicting empathic similarity, where our approach is also effective.]{\includegraphics[width=1\linewidth]{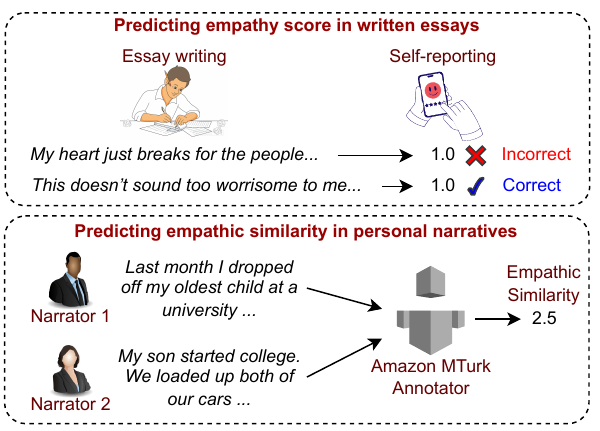}\label{fig:noise-removal-data}}
    \hfill
    \subfloat[\textbf{Left}: label noise injected (labels are shifted by 3.0) in 30\% samples, most of which are mispredicted by the model. \textit{Right}: predicted uncertainty is higher on incorrect predictions, and uncertainties correlate with noise.]{\includegraphics[width=1\linewidth]{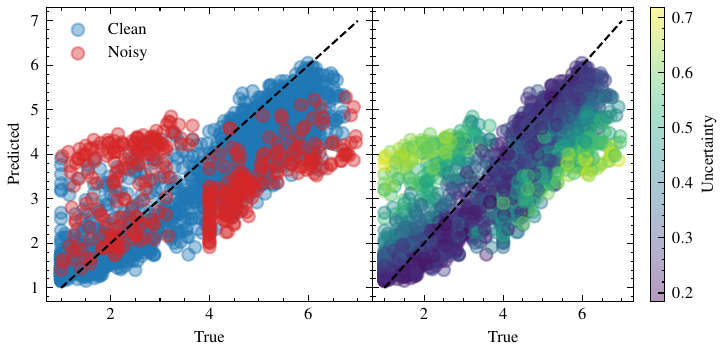}\label{fig:noise-removal-unc}}
    \caption{\textbf{(a)} Demonstration of empathy regression tasks and \textbf{(b)} Uncertainty based on our proposed penalty loss capturing label noise by ensuring higher uncertainty estimation on noisy samples and lower on clean samples.}
    \label{fig:noise-removal}
\end{figure}

\section{Introduction}
Empathy is an involuntary, vicarious reaction to another person's feelings or circumstances \citep{hoffman1978toward}, which drives supportive communication across various human-human \citep{hasan2025empathy} and human-machine \citep{banarjee2020z} interactions. Automatic empathy regression from textual content has gained traction over the past few years \citep{hasan2025empathy}. In most empathy regression studies, the \emph{``regression''} part refers to estimating the response (i.e., an aggregated continuous value) of empathy measurement scales or questionnaires that would otherwise be used for a manual assessment of empathy. Over many years, social psychology literature has developed various such empathy measurement scales. One such scale is developed by \citet{batson1987distress}, which measures empathy in terms of six dimensions: sympathetic, moved, compassionate, tender, warm and soft-hearted. Being a subjective phenomenon and further to conform to the definitions of empathy developed in social psychology literature, training labels for automated empathy detection systems often come from self-assessed empathy labels \citep{hasan2025empathy}. The de facto method of creating such datasets for empathy detection is crowdsourcing, which often results in incorrect empathy labels due to the nature of crowdsourcing \citep{sheehan2018crowdsourcing,hasan2024llm-gem}. While circumventing this issue of label noise, building a robust empathy detection system is a long-standing challenge.

\IEEEpubidadjcol

As shown in \Cref{fig:noise-removal-data}, the same empathy score (a score of 1.0 here means the lowest empathy) can be self-reported for essays with clearly different empathic content, resulting in incorrect or misleading labels. Such noise may confuse regression models, as they are penalised for producing predictions aligned with textual meaning when the corresponding label is unreliable. Without explicitly accounting for this discrepancy, models risk overfitting to noisy supervision and producing incorrect empathy scores. To address this, our method estimates uncertainty to capture samples with label noise (\Cref{fig:noise-removal-unc}).

Robust learning with noisy labels has been extensively studied in the literature for classification problems, but it is rarely explored for regression problems \citep{wang2022noisy}. Most robust learning with noisy label algorithms, e.g., \citet{zheng2020error,ye2021learning}, leverage class probabilities. Regression and classification modelling are fundamentally different due to the type of prediction (class probability versus real numbers) and subsequently the type of cost function (cross-entropy versus error-based loss formulation). Due to the above conflicts in modelling, these robust learning algorithms designed for classification do not apply to regression problems \citep{wang2022noisy,dai2023semi}. Our proposed method, UPLME, is specifically designed for textual regression problems and has been demonstrated on the emerging topic of empathy regression.

We model label noise as heteroscedastic Uncertainty Quantification (UQ). UQ, an established concept, is broadly categorised into epistemic and aleatoric uncertainty \citep{sheitz2022pitfalls,zulfiqar2025uncertainty}. Epistemic uncertainty reflects uncertainty in the model parameters due to limited data, while aleatoric uncertainty captures the inherent noise in the data itself \citep{sheitz2022pitfalls}. Aleatoric uncertainty is further divided into homoscedastic (constant across inputs) and heteroscedastic (input-dependent) uncertainty. Since label noise, by definition, varies across input samples, we capture label noise as heteroscedastic UQ. Following recent evidence in the UQ literature on dissolving the dichotomy of epistemic and aleatoric uncertainty, we model the \emph{source} of uncertainty, more specifically, annotation-related uncertainty \citep{kirchhof2025reexamining}.

To ensure a higher uncertainty score on noisy labels and a lower uncertainty score on clean labels (\Cref{fig:noise-removal}), we propose a novel penalty loss that penalises degenerate uncertainty scores (i.e., low uncertainty on noisy samples and high uncertainty on clean samples). Theoretically, empathy is dependent on how similar the pair of input texts for which empathy is being predicted, so we further propose a novel alignment loss to enforce the similarity on the representation space of the input pairs. 


UQ itself is a challenging task due to the lack of any ground truth uncertainty score for supervision \citep{shukla2025towards}. Like robust learning with noisy labels, UQ has also been less explored in regression domains due to the challenges of the output label space being continuous rather than discrete class probabilities \citep{wang2022noisy}. We propose a probabilistic fine-tuning of Pre-trained Language Models (PLMs) to estimate uncertainty using the concept of Bayesian Neural Networks (BNNs). Our key contributions include:
\begin{enumerate}[nosep]
    \item A novel penalty loss component to penalise degenerate uncertainty quantification related to label noise.
    \item A novel alignment loss component to enforce a theoretical concept of empathy.
    \item Uncertainty-aware probabilistic fine-tuning of PLM to capture label noise.
    \item State-of-the-art empathy detection performance on two public benchmarks.
\end{enumerate}


\section{Related Work \& Our Novelty}
\subsection{Empathy Detection}
There has been a growing number of empathy detection studies over the past decade, with a significantly higher number of works on Natural Language Processing (NLP) types \citep{hasan2025empathy}. We refer interested readers to a recent systematic literature review on empathy detection \citep{hasan2025empathy}. 

Task formulations for empathy detection in NLP include detecting empathy in individuals through their written essays \citep{barriere2022wassa,giorgi2024findings}, as well as empathic similarity between two narrative stories \citep{shen2023modeling}, among other scenarios. From a methodological point of view, almost all of these studies fine-tune Pre-trained Language Models (PLM), especially RoBERTa, as the backbone model \citep{hasan2025empathy}. \citet{hasan2024llm-gem} addressed label noise in empathy detection as a \emph{data-centric} approach. They re-annotated crowdsourced, noisy labels using a Large Language Model (LLM) and then selected an appropriate label between the two for each sample. Their empathy prediction model remained the same as those commonly found in the literature -- fine-tuning the RoBERTa PLM. In contrast to existing empathy detection methods in the literature, our proposed model is probabilistic, predicting both empathy and uncertainty on each prediction with two novel loss components. Our method does not require the use of any LLM and still outperforms the prior work \citep{hasan2024llm-gem} on addressing label noise in empathy detection.

\subsection{Robust Regression with Noisy Label}
Among the few works on robust regression with noisy labels, \citet{OUYANG2021ebod} proposed an ensemble-based outlier detection method by combining seven established algorithms. Since different detectors can capture different aspects of noisy data, such aggregation was able to identify the most obvious outliers agreed by the ensemble. However, combining seven algorithms increases computational complexity, and importantly, their method is demonstrated only on numerical regression tasks. In the textual regression domain, \citet{wang2022noisy} evaluated several external regularisation strategies (e.g., dropout and weight decay) and proposed an iterative algorithm to identify noisy labels. Unlike this line of work, we propose probabilistic fine-tuning of pre-trained language models that automatically downweights the contribution of noisy labels during training.

\subsection{Uncertainty Quantification}

\citet{shukla2025towards} introduced a self-supervised approach to estimate label uncertainty using covariance, where pseudo-labels are derived through a neighbourhood-based heuristic. To compare multivariate normal distributions, they reported that the 2-Wasserstein distance performs better than the KL divergence.
Recently, \citet{zulfiqar2025uncertainty} uses Monte Carlo dropout (MCD) to capture uncertainty and a Gaussian mixture model to cluster high-confidence pseudo labels for positive and unlabelled learning problems.
\citet{liang2021r-drop} proposed R-Drop, a regularisation method that minimises the divergence between outputs from two forward passes with dropout. This technique treats the two passes as stochastic sub-models within a single network. Their final loss combines negative log-likelihood with bidirectional KL divergence between the output distributions. However, R-Drop uses softmax probabilities, which makes it unsuitable for regression tasks.

\citet{wang2022uncertainty} explored UQ in textual regression settings, where they used a threshold, as a hyperparameter, on the predicted uncertainty to filter less confident samples. Contrary to their work, our method does not require any such thresholds. Finally, \citet{dai2023semi} proposed UCVME, a semi-supervised learning framework based on variational model ensembling and MCD as a BNN approach. UC in the acronym UCVME stands for uncertainty consistency, as their framework imposes consistency on the estimated uncertainties of two separate models. Their UQ framework is designed and demonstrated for computer vision problems with ResNet as the backbone model. In contrast, our method is designed and demonstrated for textual regression tasks with PLM as the backbone. It does not require two separate models or consistency between models, yet it outperforms their approach.

\section{Method}

\subsection{Problem Formulation}
Let $x_1$ and $x_2$ be a pair of input texts and $y\in\mathbb{R}$ the corresponding empathy score. In the EmpStories dataset \citep{shen2023modeling}, $y$ measures empathic similarity between two story texts. In the NewsEmp dataset \citep{buechel2018modeling}, it refers to the degree of empathy expressed in essays written by study participants after reading newspaper articles.

We focus on \emph{state} empathy in written narratives, which depends on both source and target contexts \citep{hasan2025are}. Accordingly, we adopt a \emph{cross‑encoder} formulation: we form the input sequence
\begin{equation}\label{eqn:concat}
   z = \bigl[\mathtt{CLS},x_1,\mathtt{SEP},\mathtt{SEP},x_2,\mathtt{SEP}\bigr],
\end{equation}
tokenise it jointly, and feed it into a shared encoder $\mathcal{F}_\theta$\footnote{Note that while RoBERTa tokeniser has this formulation of two SEP tokens in separating $x_1$ and $x_2$, DeBERTa and ModernBERT tokenisers assign only one SEP token in separating the inputs.}. While a bi-encoder (two separate encoders) is possible, the cross-encoder can directly capture interactions between $x_1$ and $x_2$.

\subsection{Uncertainty‑Aware Probabilistic Modelling}
To capture both the predicted empathy score and the input-dependent uncertainty, we extend a PLM with parallel regression heads. This design enables the model to jointly learn the target value and its corresponding variance, which is the foundation of our probabilistic framework. Mathematically, we augment $\mathcal{F}_\theta$ (e.g., RoBERTa, DeBERTa or ModernBERT) with two parallel regression heads:
\begin{align}\label{eqn:forward-pass}
\begin{split}
   h = \mathcal{F}_\theta(z), 
   \quad &
   \hat y = w_m^\top h + b_m,\\
   v = w_v^\top h + b_v,
   \quad &
   \sigma^2 = \mathrm{softplus}(v) + \varepsilon,
\end{split}
\end{align}
where $\hat y$ is the predicted mean, $\sigma^2$ is the aleatoric variance (clamped by $\varepsilon=10^{-8}$ for stability), and $h$ denotes the pooled sequence representation produced by the encoder, obtained by using the standard RoBERTa pooler (a linear layer followed by Tanh activation function) for RoBERTa and standard CLS pooling for other PLMs.

\subsection{Variational Model Ensembling}
We capture epistemic uncertainty using Monte Carlo dropout (MCD), which provides a variational Bayesian approximation without requiring a fully Bayesian neural network. In MCD, dropout is activated at inference time to simulate sampling from a posterior over models \citep{dai2023semi}. In particular, UPLME involves $T$ stochastic forward passes through the probabilistic language model $\mathcal{F}_\theta$ during training, validation and test phases. We collect the set $\{\hat y_i^{(t)},\sigma_i^{2,(t)},\,h_i^{(t)}\}_{t=1}^T$, representing the predicted empathy scores, variances and intermediate representations from each stochastic pass. We then compute the ensemble means:
\begin{align}\label{eqn:ensemble}
\begin{split}
  \bar y_i = \frac{1}{T} \sum_{t=1}^T \hat y_i^{(t)},
  &\quad
  \overline{\sigma^2}_i = \frac{1}{T} \sum_{t=1}^T \sigma_i^{2,(t)},\\
  \bar h_i& = \frac{1}{T} \sum_{t=1}^T h_i^{(t)}.
\end{split}
\end{align}

We use a single probabilistic PLM, which has three loss components, as described in the following subsections.

\begin{figure*}[t!]
    \centering
    \includegraphics[width=1\linewidth]{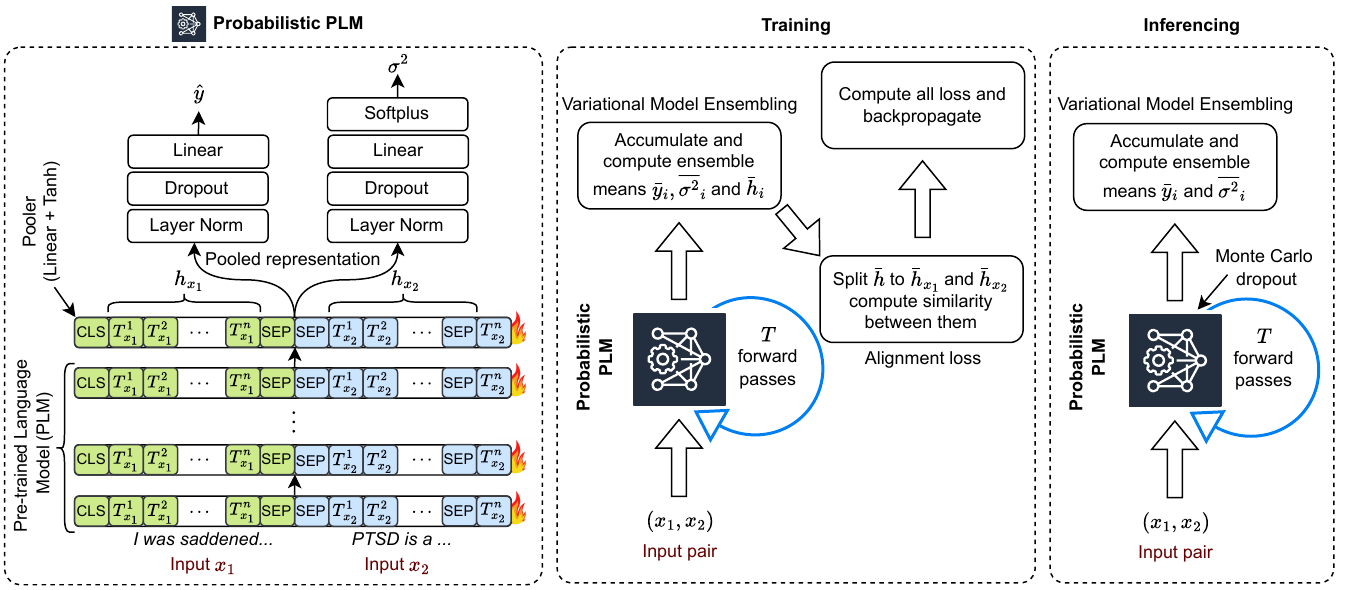}
    \caption{Overview of our proposed UPLME framework. Input pairs are first concatenated and fed to a probabilistic Pre-trained Language Model (PLM) that predicts both empathy scores and heteroscedastic uncertainty. Uncertainty quantification is stabilised through variational model ensembling that includes multiple forward passes through the same model.}
    \label{fig:overview}
\end{figure*}

\begin{algorithm*}[!t]
\caption{UPLME Training and Evaluation}\label{alg}
\begin{algorithmic}[1]
\Require 
    Batch $\mathcal{B}=\{(x_1^{(i)},x_2^{(i)},y^{(i)})\}_{i=1}^N$,
    encoder $\mathcal{F}_\theta$,
    heads $(w_m,b_m),(w_v,b_v)$,
    hyperparameters $\alpha,\lambda_1,\lambda_2$,
    passes $T$
\Ensure Total loss $\mathcal{L}_{\mathrm{total}}$

\LComment{Training using Backpropagation}
\For{each $(x_1,x_2,y)\in\mathcal{B}$}
    \State Initialise $\displaystyle \sum\hat y\gets0,\;\sum\sigma^2\gets0,\;\sum h_z\gets0$ \label{start-ensemble-loop}
    \For{$t\gets 1$ \textbf{to} $T$}
        \State Concatenate $x_1$ and $x_2$ using \Cref{eqn:concat}
        \State Forward pass through the probabilistic PLM using \Cref{eqn:forward-pass}
        \State Accumulate: 
        $\sum\hat y\!+\!=\hat y^{(t)},\;
         \sum\sigma^2\!+\!=\sigma^{2,(t)},\;
         \sum h_z\!+\!=h_z^{(t)}$
    \EndFor
    \State Compute ensemble means using \Cref{eqn:ensemble} \label{end-ensemble-loop}
    \State Compute $\mathcal{L}_{\beta\mathrm{-nll}}$ using \Cref{eqn:nll}
    \State Compute $\mathcal{L}_\mathrm{pen}$ using \Cref{eqn:pen}
    \State Split $h_z$ into $h_{x_1},h_{x_2}$, and calculate similarity using \Cref{eqn:split-h}
    \State Compute $\mathcal{L}_\mathrm{align}$ using \Cref{eqn:align}
    \State Calculate $\mathcal{L}_\mathrm{total}$ using \Cref{eqn:total-loss} and accumulate for backproagation.
\EndFor

\LComment{Evaluation}
\For{each $(x_1,x_2)$ in validation/test set}
    \State Activate dropout in $\mathcal{F}_\theta$ \Comment{Monte Carlo Dropout (MCD)}
    \State Repeat Line \ref{start-ensemble-loop} to Line \ref{end-ensemble-loop} to compute $\bar y,\overline{\sigma^2}$
    \State Use $\bar y,\overline{\sigma^2}$ for downstream metrics
\EndFor
\end{algorithmic}
\end{algorithm*}

\subsubsection{Negative Log-Likelihood}
A Negative Log-Likelihood (NLL) objective can adjust the influence of samples according to the variance parameters, which is equivalent to the estimated noise. Following \citet{sheitz2022pitfalls}, we use the improved version of it, a $\beta$-NLL loss, which down‑weights high‑variance points as:
\begin{equation}\label{eqn:nll}
  \mathcal{L}_{\beta\text{-nll}}=\frac{1}{2}\sum_{i=1}^N\overline{\sigma_i^{2}}^\beta\left(\log\overline{\sigma^2}+\frac{(y_i-\bar y_i)^2}{\overline{\sigma_i^2}}\right),
\end{equation}
where $\beta>0$ controls the degree of variance rescaling, which is set to $0.5$, following \citet{sheitz2022pitfalls}.

\subsubsection{Variance Penalty}
To discourage degenerate variance predictions (i.e., high uncertainty at low error and low uncertainty at high error), we apply an error-weighted penalty to the predicted variance. Specifically, we scale each variance by an exponentially decaying function of its squared error, and take the L2 norm across the batch:
\begin{equation}\label{eqn:pen}
    \mathcal{L}_\mathrm{pen}=\frac{1}{N}\left\|e^{-\alpha\left(y-\bar{y}\right)^2}\times\overline{\sigma^2}\right\|_2,
\end{equation}
where $\alpha>0$ modulates how sharply low-error points are penalised for high uncertainty. This formulation encourages the model to reserve high variance for genuinely uncertain samples, and therefore improves aleatoric UQ.

\subsubsection{Alignment Loss} 
The degree of alignment between two texts (e.g., a news article and its corresponding essay) should reflect the degree of empathy. To capture this, we compute the \emph{cosine similarity} between their token representations. 

Given the input sequence defined in \Cref{eqn:concat}, the encoder produces a contextualised representation $\bar h_z \in \mathbb{R}^{L \times d}$ for the full token sequence $z$. We extract individual representations $\bar h_{x_1}$ and $\bar h_{x_2}$ by splitting $\bar h_z$ at the SEP tokens and averaging the token embeddings for each segment:
\begin{align}\label{eqn:split-h}
\begin{split}
  \bar h_{x_1} = \frac1{|T_{x_1}|}\sum_{t\in T_{x_1}}\bar h_t,
  \quad &
  \bar h_{x_2} = \frac1{|T_{x_2}|}\sum_{t\in T_{x_2}}\bar h_t,\\
  s = \cos(\bar h_{x_1}&,\,\bar h_{x_2}),
\end{split}
\end{align}
where $T_{x_1}$ and $T_{x_2}$ index the token positions of $x_1$ and $x_2$ within $z$. The similarity score $s \in [-1, 1]$ captures the degree of alignment between the two texts. To use this as a learning signal, we first rescale the original empathy labels (e.g., $[1, 7]$) to $[-1, 1]$) to match the scale of $s$. Let $y'$ denote the transformed label. We then define the alignment loss as
\begin{equation}\label{eqn:align}
  \mathcal{L}_{\mathrm{align}}
  = \frac{1}{N}\sum_{i=1}^N (s_i - y_i')^2.
\end{equation}
This objective encourages the model to align the similarity between two texts with their associated empathy scores.

The overall loss for \emph{a single probabilistic model} is, therefore:
\begin{equation}\label{eqn:total-loss}
  \mathcal{L}=\mathcal{L}_{\beta\text{-nll}}+\lambda_1\mathcal{L}_{\mathrm{pen}}+\lambda_2 \mathcal{L}_{\mathrm{align}},
\end{equation}
with weights $\lambda_{1,2}$ to be tuned as hyperparameters.

\Cref{fig:overview} provides an overview of the complete UPLME framework. On the left, the probabilistic PLM is shown: input pairs $(x_1, x_2)$ are concatenated and jointly encoded, with a pooled representation passed through two regression heads to predict the empathy score $\hat{y}$ and heteroscedastic variance $\sigma^2$, respectively. During the training phase (middle), UPLME performs $T$ stochastic forward passes and accumulates the predicted means, the variances, and the hidden representations for loss computations. The loss function consists of three components: the likelihood-based objective, the variance penalty, and the alignment loss between the representations of $x_1$ and $x_2$. During inference (right), the same procedure is repeated with MCD to approximate posterior sampling, and ensemble averages of the predictions are used as final outputs. Overall training and evaluation steps of UPLME are shown in \Cref{alg}.

\subsection{UPLME vs Others}
UPLME differs substantially from prior uncertainty-aware regression approaches in both design and training. Unlike UCVME \citep{dai2023semi}, which co-trains two ResNet-based models with an uncertainty consistency loss in a semi-supervised setting for computer vision tasks, UPLME utilises a single PLM backbone for textual regression and achieves heteroscedastic uncertainty estimation without requiring dual models or explicit consistency constraints. Moreover, previous methods for robust regression under label noise often rely on external heuristics to filter or correct noisy data -- for example, combining multiple outlier detectors into an ensemble \citep{OUYANG2021ebod} or iteratively removing suspected noisy labels in textual datasets \citep{wang2022noisy} -- and some require manual uncertainty thresholds to discard low-confidence predictions \citep{wang2022uncertainty}. In contrast, UPLME directly learns a probabilistic output distribution and treats label noise as input-dependent (heteroscedastic) uncertainty, automatically down-weighting the influence of noisy examples via a $\beta-$NLL loss term instead of ad-hoc filtering. In addition to excelling in the UQ task, UPLME introduces two novel loss components: a variance penalty that discourages miscalibrated uncertainty estimates, and an alignment loss that reflects the semantic similarity of input pairs.

In terms of empathy detection workflows, prior works (e.g., \citep{giorgi2024findings,hasan2023curtin,hasan2024llm-gem} (\Cref{tab:sota-result}, to be presented later) primarily fine-tuned PLMs with their default mean squared error objective function. In contrast, UPLME comprises penalty loss, alignment loss, and variational model ensembling within a UQ framework.

\section{Experiments and Results}
\subsection{Dataset \& Preprocessing}
We validate UPLME across three datasets: two different datasets from the NewsEmp series \citep{tafreshi2021wassa,giorgi2024findings} and the EmpStories dataset \citep{shen2023modeling}. \Cref{tab:data-stat} presents statistics of the three datasets.

\begin{table}[t!]
    \centering
    \caption{Statistics of the datasets used to validate UPLME.}
    \label{tab:data-stat}
    \begin{tabular}{l*4c} \toprule
        \textbf{Name} & \textbf{\# Train} & \textbf{\# Validation} & \textbf{\# Test}  & \textbf{\# Total} \\ \midrule
        NewsEmp21 \citep{tafreshi2021wassa} & 1,860 & 270 & 525 & 2,655 \\
        NewsEmp24 \citep{giorgi2024findings} & 1,000 & 63 & 83 & 1,146 \\ 
        EmpStories \citep{shen2023modeling} & 1,500 & 100 & 400 & 4,000 \\ \bottomrule
    \end{tabular}
\end{table}

\subsubsection{NewsEmp Datasets} 
\citet{buechel2018modeling} pioneered the study of how people empathise with stories of suffering related to different entities (e.g., humans, animals and environment) portrayed in newspaper articles. After reading the articles, crowdsourced participants self-assessed their \emph{state} empathy through Batson's empathy scale \citep{batson1987distress} and subsequently wrote 300--800 character essays about their thoughts and feelings towards the news. The empathy label in this dataset is a continuous score ranging from $1.0$ to $7.0$, which is derived from the essay writers' self-assessments. Multiple versions of this original dataset have been utilised in empathy detection challenges under the ``Workshop on Computational Approaches to Subjectivity, Sentiment \& Social Media Analysis (WASSA)'' \citep{giorgi2024findings}. Among these, the 2021 and 2024 releases, hereinafter referred to as the NewsEmp21 and NewsEmp24 datasets, are distinct datasets with no overlapping samples and are therefore used in this work. 
However, some samples from the first 2018 release \citep{buechel2018modeling} and some samples from the 2023 release \citep{omitaomu2022empathic} overlap with the NewsEmp21 and NewsEmp24 datasets, respectively, and are therefore not used in this work.


Our motivation behind choosing the NewsEmp datasets is two-fold. First, these datasets are the most widely used public benchmarks across all empathy benchmarks in the literature \citep{hasan2025empathy}, allowing us to compare our method's performance across a broader range of works. Second, prior work \citep{hasan2024llm-gem} demonstrated label noise in these datasets, supported by evidence of noise and prior literature suggesting that crowdsourcing often results in incorrect data \citep{sheehan2018crowdsourcing,jia2017using,huang2012detecting,obrochta2021anomalous}. Having such label noise, the NewsEmp datasets fit well with the goal of UPLME -- robust empathy regression under the label noise problem.

The news articles in the NewsEmp datasets are long sequences, with a maximum length of 20,047 characters. To accommodate these articles in the typical PLMs' (e.g., RoBERTa) context length, we use the summarised articles reported in \citep{hasan2023curtin}, which makes the maximum length only 987 characters.

Following \citet{giorgi2024findings}'s pioneering work on the NewsEmp24 dataset, we train models using the combined training sets of multiple releases. Specifically, we combine the training set of the latest 2024 release with the training and development sets of its 2021 release. We further augment the training data using WordNet-based synonym substitution, replacing 10\% of the words in each training sample. This augmentation doubles the size of the training set and introduces lexical diversity while preserving original semantics. This resulted in a total of 6260 training samples, while the validation set (63 samples) remains the same as the original 2024 release \citep{giorgi2024findings}. We evaluate the same trained model on both the NewsEmp21 and NewsEmp24 test sets, which contain 525 and 83 samples, respectively.

\subsubsection{EmpStories Dataset}
\citet{shen2023modeling} recently released pairs of narrative stories and corresponding empathic similarity. Two human annotators rate empathic similarity on a 4-point Likert scale, which is later averaged, resulting in the final empathy labels as $\{1.0, 1.5, \ldots, 3.5, 4.0\}$. 

Although we do not suggest any presence of label noise in the EmpStories dataset, we use this dataset to demonstrate the effectiveness of UPLME on a different type of dataset apart from NewsEmp. Key difference between the NewsEmp and the EmpStories includes: (1) the former (NewsEmp) includes newspaper articles and written essays, while the latter (EmpStories) includes a pair of similar personal narrative stories, and (2) the former is annotated through \emph{self-assessment} by the essay writers themselves, while the latter is annotated by two third-party annotators.



\subsection{Evaluation Protocol}
Pearson Correlation Coefficient (PCC) is the metric adopted by the WASSA empathy detection challenge for the NewsEmp datasets. While PCC capture the linear relationship between the prediction and the ground truth, it fails to capture the magnitude of prediction error. Therefore, in addition to PCC, we report Concordance Correlation Coefficient (CCC) and Root Mean Square Error (RMSE). CCC takes into account both the linear relationship and the magnitude of the error, whereas RMSE solely captures the magnitude of the error.

To evaluate the quality UQ, we adopt all three UQ metrics proposed by \citet{wang2022uncertainty}: Calibration error (CAL), Sharpness (SHP) and Negative Log-Probability Density (NLPD).

The training, validation and test splits are already predefined in all three datasets we used. Note that the ground-truth labels for the NewsEmp21 test set are not publicly released; evaluation is only available through the CodaLab challenge platform\footnote{\url{https://codalab.lisn.upsaclay.fr/competitions/834}, noting that the same dataset has been used both in the WASSA 2021 \citep{tafreshi2021wassa} and the 2022 \citep{barriere2022wassa} challenges.}. We run each model using three different initialisations, set by random seeds of 0, 42, and 100, and compare based on median and peak (i.e., the best) scores of the models across these three seeds.

\begin{table}[t!]
    \centering
    \caption{Comparison of our proposed model's performance with the performance \emph{reported in the literature}. A single RoBERTa model under the UPLME framework, optimised for the NewsEmp24 validation set, is tested on both the NewsEmp21 and NewsEmp24 test sets.}
    \label{tab:sota-result}
    \resizebox{\columnwidth}{!}{%
    \begin{tabular}{@{}*3lc@{}} \toprule
        \textbf{Reference} & \textbf{Venue \& Year} & \textbf{Model} & \textbf{PCC $\uparrow$} \\ \midrule
        \texttt{\textit{NewsEmp21}} \\
        \citet{butala2021team} & ACL WASSA 2021 & BERT-MLP & 0.358 \\
        \citet{vasava2022transformer} & ACL WASSA 2022 & RoBERTa-MLP & 0.470 \\
        \citet{ghosh2022team} & ACL WASSA 2022 & BERT-MLP & 0.479 \\
        \citet{qian2022surrey} & ACL WASSA 2022 & RoBERTa & 0.504 \\
        \citet{hasan2024llm-gem} & EACL 2024 & RoBERTa-MLP & 0.505 \\
        \citet{vettigli2021empna} & EACL WASSA 2021 & LR & 0.516 \\
        \citet{kulkarni2021pvg} & EACL WASSA 2021 & RoBERTa-MLP & 0.517 \\
        \citet{lahnala2022caisa} & ACL WASSA 2022 & RoBERTa & 0.524 \\
        \citet{chen2022iucl} & ACL WASSA 2022 & RoBERTa & 0.537 \\
        \citet{plaza2022empathy} & ACL WASSA 2022 & RoBERTa & 0.541 \\
        \citet{mundra2021wassa} & EACL WASSA 2021 & ELECTRA + RoBERTa & 0.558 \\
        \textbf{Ours} & -- & \textbf{UPLME (RoBERTa)} & \textbf{0.580} \\ \midrule
        
        \texttt{\textit{NewsEmp24}} \\
        \citet{numanoglu2024empathify} & ACL WASSA 2024 & BERT & 0.290 \\
        \citet{chevi2024daisy} & ACL WASSA 2024 & MLP & 0.345 \\ 
        \citet{frick2024fraunhofer} & ACL WASSA 2024 & RoBERTa & 0.375 \\
        \citet{pereira-etal-2024-context} & ACL WASSA 2024 & \textit{Not reported} & 0.390 \\
        \citet{li2024chinchunmei} & ACL WASSA 2024 & Llama 3 8B & 0.474 \\
        \citet{giorgi2024findings} & ACL WASSA 2024 & RoBERTa & 0.629 \\
        \textbf{Ours} & -- & \textbf{UPLME (RoBERTa)} & \textbf{0.634} \\
        \bottomrule
    \end{tabular}%
    }
\end{table}

\subsection{Implementation Details}
UPLME was primarily implemented using PyTorch 2.2.0 with ROCm 5.7.3 in Python 3.10 and also tested on PyTorch 2.7.1 with ROCm 6.3.3 in Python 3.12.3. Deterministic training and evaluation are ensured using the PyTorch Lightning package. All experiments were conducted on a single AMD Instinct™ MI250X GPU (64 GB) running on SUSE Linux Enterprise Server 15 SP5.

Pre-trained checkpoints of RoBERTa (version: \texttt{roberta-base}) \citep{liu2019roberta}, DeBERTa (version: \texttt{microsoft/deberta-v3-base}) \citep{he2023debertav3}, and ModernBERT (version: \texttt{answerdotai/ModernBERT-base}) \citep{benjamin2024smarter} are accessed from HuggingFace. We use the TextAttack package \citep{morris2020textattack} for training data augmentation.

Automated tuning of loss weights was conducted using the Optuna hyperparameter tuning framework \citep{akiba2019optuna}, with the median pruner (warmup steps $= 5$) and the default ``tree-structured Parzen estimator'' as the sampler. Tuning was conducted between \{1.0, 1.5\} for $\alpha$ and [0, 50] with a step of 0.1 for both $\lambda_1$ and $\lambda_2$, with the objective of selecting the final set of parameters that yields the best validation CCC score. We tuned $\alpha, \lambda_1$ and $\lambda_2$ of UPLME and $w$ (factor weighing the contribution of two models' output consistency) of UCVME \citep{dai2023semi}. 

Other hyperparameters were kept fixed throughout the experiment, with most of them being selected from the literature. For example, following the pioneering work \citep{liu2019roberta} releasing RoBERTa, we set $\beta_1=0.9$, $\beta_2=0.98$, $\epsilon=$\num{1e-6} and weight decay $=0.1$ of the \texttt{AdamW} optimiser and train (fine-tune) with a learning rate of \num{3e-5}. DeBERTa was fine-tuned with a learning rate of \num{3e-5} and weight decay of 0.01, as reported in \citep{he2023debertav3}. ModernBERT was fine-tuned with the same learning rate and weight decay as RoBERTa. Training was conducted for a maximum of 6,000 steps with a batch size of 16 for both training and evaluation. We used a linear learning rate scheduler with a warmup period of the first 3\% of the training steps.
We employ a delayed-start early-stopping strategy, which ensures a minimum of 5 epochs of training, followed by early stopping based on the validation CCC score.

\subsection{Main Results}
We benchmark the performance of UPLME against the State-Of-The-Art (SOTA) performance \emph{reported in the literature} (\Cref{tab:sota-result}) and also against recent baseline methods implemented by us (\Cref{tab:sota-methods}). Since RoBERTa is the most commonly used backbone for empathy regression across the datasets we experimented \citep{hasan2025empathy}, we primarily report results using RoBERTa unless specified otherwise.

\begin{table*}[!t]
    \centering
    \caption{UPLME versus our implementation of baseline methods, trained and evaluated on the NewsEmp24 and the EmpStories datasets (These metrics on the NewsEmp21 dataset were not calculated due to the unavailability of ground truth test labels). UCVME is a recent uncertainty-aware variational ensembling-based approach.}
    \label{tab:sota-methods}
    \begin{threeparttable}
    \resizebox{\linewidth}{!}{%
    \begin{tabular}{@{}*1l*7c@{}} \toprule
        \textbf{Framework} & \textbf{PCC $\uparrow$} & \textbf{CCC $\uparrow$} & \textbf{SCC $\uparrow$} & \textbf{RMSE $\downarrow$} & \textbf{CAL $\downarrow$} & \textbf{SHP $\downarrow$} & \textbf{NLPD $\downarrow$} \\ \midrule
        \texttt{\textit{NewsEmp24}} & \\
        PLM-MLP & $0.554(0.569)$ & $0.476(0.499)$ & $0.453(0.488)$ & $1.384(1.368)$ \\
        UCVME \citep{dai2023semi} & $0.587(0.589)$ & $0.510(0.530)$ & $0.493(0.496)$ & $1.337(1.317)$ & $0.571(0.582)$ & $\textbf{0.001(0.001)}^{***}$ & $2085042.875(2449522.5)$ \\
        UPLME (Ours) & $\textbf{0.623(0.634)}$ & $\textbf{0.523(0.534)}$ & $\textbf{0.503(0.53)}$ & $\textbf{1.322(1.313)}$ & $\textbf{0.376(0.402)}^{**}$ & $0.268(0.291)$ & $\textbf{10.775(14.718)}^{***}$\\ \midrule
        \texttt{\textit{EmpStories}} \\
        PLM-MLP & $0.180(0.215)$ & $0.113(0.150)$ & $0.168(0.212)$ & $0.755(0.745)$ \\
        UCVME \citep{dai2023semi} & $0.239(0.250)$ & $0.180(0.189)$ & $0.244(0.244)$ & $0.748(0.741)$ & $0.534(0.566)$ & $\textbf{0.002(0.003)}^{***}$ & $94886.258(142711.703)$ \\
        UPLME (Ours) & $\textbf{0.256(0.301)}$ & $\textbf{0.217(0.237)}$ & $\textbf{0.250(0.292)}$ & $\textbf{0.747(0.722)}$ & $\textbf{0.262(0.289)}^{***}$ & $0.186(0.232)$ & $\textbf{6.759(7.477)}^{*}$ \\
    \bottomrule
    \end{tabular}}
    \begin{tablenotes}
        \item Reported metrics are in \textit{median(peak)} format, calculated over three random initialisations. 
        \item \textbf{Boldface} scores indicate the best scores. 
        \item Asterisk (*) refer to the statistical significance level from an independent t-test between UCVME and UPLME (* means $0.01 < p\text{-value} \leq 0.05$; ** means $0.001 < p\text{-value} \leq 0.01$; *** means $0.0001 < p \text{-value}\leq 0.001$)
    \end{tablenotes}
    \end{threeparttable}
\end{table*}

\begin{table*}[!t]
    \centering
    \caption{Test set performance of UPLME with different model configurations on the NewsEmp24 and EmpStories datasets.}
    \label{tab:ablation}
    \begin{threeparttable}
    \resizebox{\linewidth}{!}{%
    \begin{tabular}{@{}*2l*7c@{}} \toprule
        \textbf{Theme\tnote{$\dagger$}} & \textbf{Configuration} & \textbf{PCC $\uparrow$} & \textbf{CCC $\uparrow$} & \textbf{SCC $\uparrow$} & \textbf{RMSE $\downarrow$} & \textbf{CAL $\downarrow$} & \textbf{SHP $\downarrow$} & \textbf{NLPD $\downarrow$} \\ \midrule
        %
        A\textsubscript{NewsEmp} & $\lambda_{1,2}=0$ & $0.580(0.597)$ & $\textbf{0.514}(0.516)$ & $0.483(0.503)$ & $\textbf{1.356(1.318)}$ & $0.362(0.393)$ & $0.270(\textbf{0.281})$ & $11.555(14.63)$ \\
           & $\lambda_{1}=1, \lambda_{2}=0$ & $\textbf{0.586(0.610)}$ & $0.512(\textbf{0.529})$ & $\textbf{0.513(0.526)}$ & $1.365(1.340)$ & $0.318(\textbf{0.323})$ & $0.310(0.318)$ & $\textbf{8.668(11.514)}$ \\
           & $\lambda_{1}=0, \lambda_{2}=1$ & $0.552(0.563)$ & $0.489(0.517)$ & $0.466(0.468)$ & $1.386(1.351)$ & $\textbf{0.292}(0.339)$ & $0.279(0.317)$ & $10.630(11.966)$\\
           & $\lambda_{1,2}=1$ & $0.551(0.579)$ & $0.486(0.499)$ & $0.461(0.487)$ & $1.404(1.39)$ & $0.369(0.388)$ & $\textbf{0.268}(0.343)$ & $12.498(16.858)$ \\ \midrule
        A\textsubscript{EmpStories} & $\lambda_{1,2}=0$ & $0.226(0.262)$ & $0.185(0.188)$ & $0.220(0.267)$ & $0.755(0.740)$ & $0.232(0.285)$ & $0.269(0.326)$ & $2.839(4.466)$ \\
           & $\lambda_{1}=1, \lambda_{2}=0$ & $\textbf{0.254}(0.272)$ & $\textbf{0.203}(0.212)$ & $\textbf{0.252}(0.270)$ & $\textbf{0.747(0.737)}$ & $0.242(0.265)$ & $\textbf{0.230(0.298)}$ & $3.784(3.955)$ \\
           & $\lambda_{1}=0, \lambda_{2}=1$ & $0.204(\textbf{0.303})$ & $0.158(\textbf{0.249})$ & $0.215(\textbf{0.302})$ & $0.784(0.782)$ & $0.208(0.274)$ & $0.311(0.353)$ & $2.021(2.596)$ \\
           & $\lambda_{1,2}=1$ & $0.231(0.239)$ & $0.197(0.197)$ & $0.225(0.228)$ & $0.759(0.741)$ & $\textbf{0.168(0.185)}$ & $0.32(0.333)$ & $\textbf{1.773(2.347)}$ \\ \midrule
        %
        B\textsubscript{NewsEmp} & $T=1$ & $0.568(\textbf{0.593})$ & $0.509(0.518)$ & $0.474(0.494)$ & $1.369(1.337)$ & $0.359(0.416)$ & $0.279(0.301)$ & $\textbf{11.948}(15.999)$ \\
          & $T=2$ & $0.556(0.559)$ & $0.493(0.503)$ & $0.479(0.481)$ & $1.376(1.364)$ & $\textbf{0.324}(0.379)$ & $0.257(0.280)$ & $13.598(15.349)$ \\
          & $T=3$ & $0.572(0.584)$ & $0.486(0.530)$ & $0.482(0.485)$ & $1.376(1.328)$ & $0.399(0.403)$ & $\textbf{0.240(0.241)}$ & $15.289(17.327)$ \\
          & $T=4$ & $\textbf{0.578}(0.584)$ & $\textbf{0.524(0.533)}$ & $\textbf{0.488(0.496)}$ & $\textbf{1.329(1.314)}$ & $0.336(\textbf{0.376})$ & $0.260(0.293)$ & $11.976(\textbf{13.819})$ \\
          & $T=5$ & $\textbf{0.578}(0.579)$ & $0.518(0.521)$ & $0.466(0.467)$ & $1.342(1.328)$ & $0.356(0.396)$ & $0.239(0.264)$ & $13.984(16.252)$ \\
          \midrule
        B\textsubscript{EmpStories} & $T=1$ & $0.283(0.283)$ & $0.235(0.237)$ & $\textbf{0.288}(0.291)$ & $\textbf{0.740}(0.735)$ & $0.243(0.268)$ & $\textbf{0.204(0.207)}$ & $5.088(\textbf{5.259})$ \\
          & $T=2$ & $0.280(0.289)$ & $0.232(0.247)$ & $0.279(0.280)$ & $0.743(0.735)$ & $0.236(0.270)$ & $0.222(0.231)$ & $4.501(5.761)$ \\
          & $T=3$ & $0.251(\textbf{0.311})$ & $0.220(\textbf{0.259})$ & $0.240(\textbf{0.302})$ & $0.763(0.725)$ & $\textbf{0.203}(0.289)$ & $0.232(0.243)$ & $4.162(7.574)$ \\
          & $T=4$ & $\textbf{0.289}(0.308)$ & $\textbf{0.245}(0.258)$ & $0.279(0.298)$ & $0.759(\textbf{0.722})$ & $0.226(\textbf{0.244})$ & $0.229(0.232)$ & $\textbf{4.153}(6.616)$ \\
          & $T=5$ & $0.263(0.269)$ & $0.220(0.233)$ & $0.248(0.284)$ & $0.751(0.736)$ & $0.264(0.274)$ & $0.210(0.220)$ & $4.930(7.666)$ \\
          \midrule
        %
        C\textsubscript{NewsEmp} & Augmentation & $\textbf{0.578(0.584)}$ & $\textbf{0.524(0.533)}$ & $\textbf{0.488(0.496)}$ & $\textbf{1.329(1.314)}$ & $0.336(0.376)$ & $\textbf{0.260(0.293)}$ & $11.976(13.819)$ \\
          & No augmentation & $0.436(0.508)$ & $0.354(0.426)$ & $0.369(0.413)$ & $1.439(1.367)$ & $\textbf{0.244(0.290)}$ & $0.489(0.547)$ & $\textbf{3.191(5.977)}$ \\ \midrule
        C\textsubscript{EmpStories} & Augmentation & $\textbf{0.289(0.308)}$ & $\textbf{0.245(0.258)}$ & $\textbf{0.279(0.298)}$ & $0.759(\textbf{0.722})$ & $\textbf{0.226(0.244)}$ & $\textbf{0.229(0.232)}$ & $4.153(6.616)$ \\
          & No augmentation & $0.249(0.301)$ & $0.210(0.229)$ & $0.269(0.296)$ & $\textbf{0.752}(0.735)$ & $\textbf{0.226}(0.246)$ & $0.235(0.281)$ & $\textbf{3.56(4.417)}$ \\ \midrule
        D\textsubscript{NewsEmp} & RoBERTa & $\textbf{0.623(0.634)}$ & $\textbf{0.523(0.534)}$ & $\textbf{0.503(0.530)}$ & $\textbf{1.322(1.313)}$ & $\textbf{0.376(0.402)}$ & $0.268(0.291)$ & $\textbf{10.775(14.718)}$ \\
            & ModernBERT & $0.527(0.562)$ & $0.467(0.502)$ & $0.467(0.482)$ & $1.446(1.367)$ & $0.386(0.461)$ & $0.257(0.267)$ & $15.505(20.052)$ \\
            & DeBERTa & $0.568(0.595)$ & $0.473(0.510)$ & $0.464(0.517)$ & $1.388(1.383)$ & $0.414(0.470)$ & $\textbf{0.227(0.255)}$ & $17.978(21.78)$ \\ \midrule
        D\textsubscript{EmpStories} & RoBERTa & $\textbf{0.289(0.308)}$ & $\textbf{0.245(0.258)}$ & $\textbf{0.279(0.298)}$ & $0.759(0.722)$ & $\textbf{0.226(0.244)}$ & $0.229(\textbf{0.232})$ & $\textbf{4.153}(6.616)$ \\
            & ModernBERT & $0.186(0.254)$ & $0.157(0.159)$ & $0.179(0.25)$ & $0.784(0.720)$ & $0.259(0.324)$ & $0.203(0.318)$ & $6.041(10.204)$ \\
            & DeBERTa & $0.240(0.309)$ & $0.211(0.248)$ & $0.228(0.286)$ & $\textbf{0.757(0.717)}$ & $0.252(0.258)$ & $\textbf{0.198}(0.234)$ & $5.783(\textbf{5.853})$ \\
    \bottomrule
    \end{tabular}%
    }
    \begin{tablenotes}
        \item[$\dagger$] A -- different loss components; B -- variational ensembling; C -- training data augmentation; D -- different PLM encoders.
        \item Reported metrics are in \textit{median(peak)} format, calculated over three random initialisations. 
        \item \textbf{Boldface} scores indicate the best scores in each theme.
    \end{tablenotes}
    \end{threeparttable}
\end{table*}

\subsubsection{Benchmarking Results}
Since the literature on the NewsEmp datasets primarily uses the PCC evaluation metric, \Cref{tab:sota-result} compares our method's performance with the performance reported in the literature in terms of the PCC score. On both datasets, UPLME outperforms all prior results.

The prior SOTA work \citep{mundra2021wassa} on the NewsEmp21 dataset employs an ensemble of ELECTRA and RoBERTa, reporting a PCC of 0.558, which is now surpassed by our proposed UPLME framework using a single RoBERTa PLM. On the NewsEmpat24 dataset, our method also provides the best PCC score, even outperforming Llama 3 LLM used in \citep{li2024chinchunmei}.

For the NewsEmp21 dataset, we do not retrain the model; instead, we use the same model that was optimised for the NewsEmp24 dataset. Even so, UPLME outperforms other studies on the NewsEmp21 dataset, which suggests that UPLME generalises well.

\subsubsection{Benchmarking Methods}
In addition to comparing our results with those reported in the literature, we also compare UPLME with two baseline methods. 
PLM-MLP is UPLME without any UQ and so uses a mean squared error loss function, similar to many methods (e.g., \citep{giorgi2024findings,hasan2023curtin,hasan2024llm-gem}) presented on \Cref{tab:sota-result}. 
UCVME is a UQ method that utilises variational model ensembling, but it requires two models to enforce consistency in the uncertainty prediction. Although originally proposed for computer vision tasks using ResNet, we apply the same modelling technique (two models and their loss functions) to two RoBERTa PLMs. 
As presented in \Cref{tab:sota-methods}, our method outperforms both PLM-MLP and UCVME in terms of regression metrics (PCC, CCC, SCC and RMSE).

In terms of UQ metrics, our method outperforms UCVME in CAL and NLPD metrics. The work \citep{wang2022uncertainty} proposing these UQ metrics noted that when a model achieves the best SHP but has a very large NLPD (as is the case here for UCVME), we should compare models based on PCC, followed by CAL and NLPD. We can, therefore, conclude that our proposed method, UPLME, outperforms UCVME in the UQ task as well.



\begin{figure*}[t!]
    \centering
    \includegraphics[width=1\linewidth]{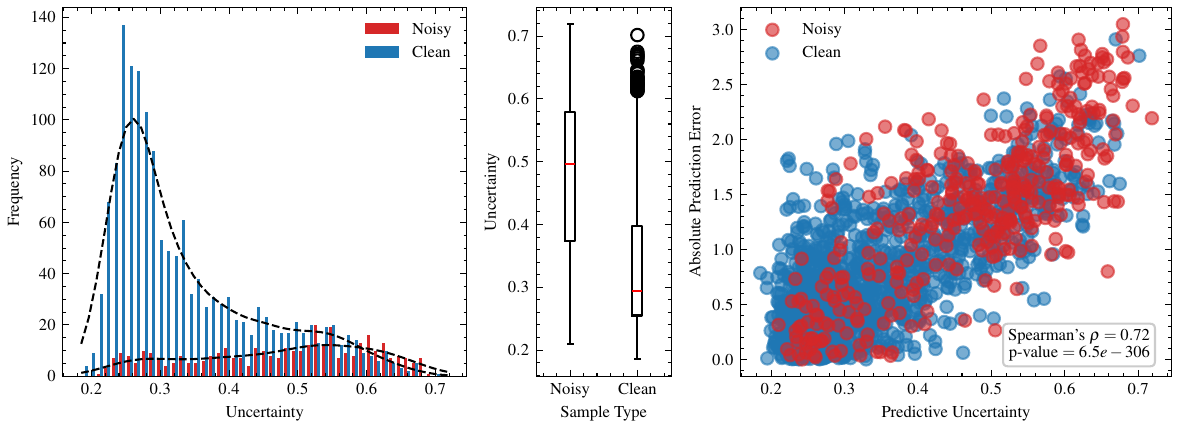}
    \caption{Our proposed penalty loss component ensures predicted uncertainty captures label noise. \textbf{Left} and \textbf{Middle}: comparison of predictive uncertainty between noisy and clean samples, showing that UPLME estimates higher uncertainty for noisy samples. \textbf{Right}: relationship between absolute prediction error and estimated uncertainty on both noisy and clean subsets, showing strong statistically-significant positive correlation (Spearman's $\rho = 0.72$).}
    \label{fig:noise-analysis}
\end{figure*}

\begin{figure}[h!]
    \centering
    \includegraphics[width=0.99\linewidth,trim={0 0 0 0.7cm},clip]{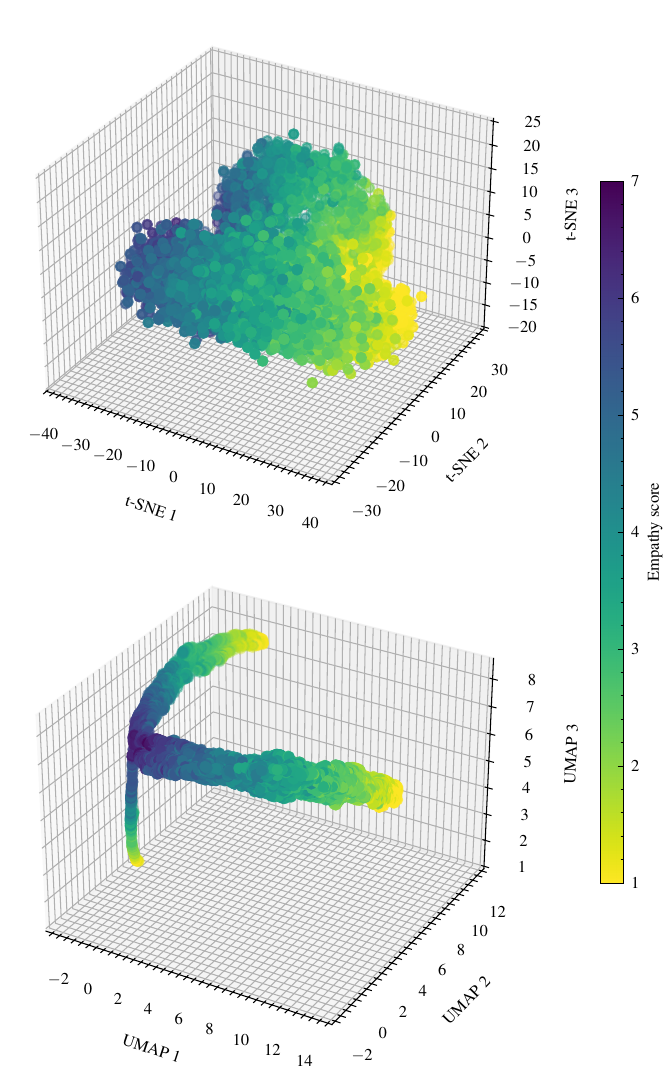}
    \caption{3D t‑SNE (left) and UMAP (right) projections of UPLME's learned representations on the NewsEmp training set. Each point corresponds to an input essay and is coloured by its ground-truth empathy score (1–7).}
    \label{fig:tsne-umap}
\end{figure}

\subsection{Ablation Study}
To better understand the contribution of each component and hyperparameters of UPLME, we analyse model performance on both NewsEmp24 and EmpStories test sets under different configurations as reported in \Cref{tab:ablation}. As demonstrated in \emph{Theme A}, when both penalty and alignment losses are removed ($\lambda_{1,2}=0$), performance drops substantially across most metrics in both datasets. Adding only the penalty loss ($\lambda_1=1, \lambda_2=0$) yields the largest improvement. The alignment loss alone ($\lambda_1=0, \lambda_2=1$ improves calibration (lower median CAL) but contributes less to empathy prediction performance in the NewsEmp dataset. A plausible explanation is that, while alignment between articles and essays is theoretically grounded, each article is paired with multiple essays that often correspond to different empathy scores, which weakens the consistency of this assumption.

\emph{Theme B} shows that, overall, variational model ensembling with $T=4$ demonstrates better performance than a single forward pass with no ensembling ($T=1$) and ensembling with $T=\{2,3,5\}$. In \emph{Theme C}, data augmentation shows performance improvements over no augmentation in both datasets.

While we primarily demonstrated the effectiveness of UPLME using the RoBERTa PLM, \Cref{tab:ablation} \emph{Theme D} shows that UPLME is independent of the specific PLM choice. RoBERTa is replaced by ModernBERT or DeBERTa, which work without any modification to the UPLME framework, suggesting that UPLME can be readily adapted to various transformer-based encoders. RoBERTa yields the highest scores, which supports the general consensus that RoBERTa has become the most popular PLM in the empathy computing literature \citep{hasan2025empathy}.

\subsection{Qualitative Analysis}
\Cref{fig:noise-analysis} evaluates whether the uncertainty scores estimated by UPLME behave as intended under controlled noise injection. For this demonstration, 30\% of the training samples’ labels are shifted by 3.0 (e.g., $1.0\rightarrow4.0$) to create a noisy subset. First, the uncertainty distribution significantly differs between clean and noisy samples: noisy samples receive higher variance estimates, while clean samples concentrate around lower uncertainty regions. Second, the predictive uncertainty strongly correlates with absolute prediction error (Spearman’s $\rho=0.72$, with a p-value of \num{6.5e-306}), which implies that the estimated uncertainty captures label noise with a very high statistical significance.

\Cref{fig:tsne-umap} examines how UPLME organises input contexts in its latent representation space. Both t-SNE and UMAP reveal a clear, continuous gradient in the embedding geometry that aligns with the ground-truth empathy scores, suggesting that UPLME has learned representations that correspond to the ordinal scale of empathy.

\subsection{Limitations}
While UPLME demonstrates robust performance, it has a few limitations. First, the variance penalty and alignment loss introduce additional hyperparameters whose optimal values may vary across datasets. Second, the use of variational ensembling increases computational cost relative to baselines with a single PLM, which may hinder scalability for larger model backbones or datasets. Nevertheless, the computational requirement is still smaller than that of LLMs or multiple ensembles of PLMs.

\section{Conclusion}
Noisy labels challenge empathy regression from self-reported annotation. This paper proposes UPLME, an uncertainty-aware probabilistic language modelling framework for empathy regression tasks from text sequences. UPLME estimates label noise-related uncertainty to down-weight the contribution from these noisy samples during training. The proposed framework includes two novel loss components, one to stabilise UQ and another to enforce similarity on the input text pairs on which empathy is being detected. Experiments across three datasets show that UPLME is effective both in providing the SOTA empathy detection performance and in improving the UQ. In future, UPLME can be extended to semi-supervised learning frameworks, where the uncertainty estimation can improve pseudo-label quality.

\section*{Ethics Statement}
This work uses publicly available empathy datasets collected through prior studies. All datasets were originally released with institutional ethics approval, and we followed their intended use for research purposes. Our method focuses on improving robustness to noisy self-reported labels and does not attempt to infer private or sensitive attributes such as demographic details. We acknowledge that automatic empathy detection may be misused, and emphasise that our framework is designed strictly for advancing methodological research in empathy and similar textual regression tasks.

\section*{Acknowledgments}
This work was supported by resources provided by the Pawsey Supercomputing Research Centre with funding from the Australian Government and the Government of Western Australia. We thank Dr Moloud Abdar, Senior Data Scientist at The University of Queensland, for sharing his opinion on uncertainty quantification during an early stage of this work.

\printbibliography


 





\end{document}